\documentclass[10pt,twocolumn,letterpaper]{article}
\usepackage[accsupp]{axessibility}
\usepackage{iccv}
\usepackage{times}
\usepackage{epsfig}
\usepackage{graphicx}
\usepackage{amsmath}
\usepackage{amssymb}
\usepackage{color}
\usepackage[accsupp]{axessibility}  
\usepackage{graphicx}
\usepackage{amsmath}
\usepackage{amssymb}
\usepackage{booktabs}
\usepackage{soul}
\usepackage{authblk}
\usepackage{multirow}
\usepackage{wrapfig}
\usepackage{array}
\newcolumntype{L}[1]{>{\raggedright\let\newline\\\arraybackslash\hspace{0pt}}m{#1}}
\newcolumntype{C}[1]{>{\centering\let\newline\\\arraybackslash\hspace{0pt}}m{#1}}
\newcolumntype{R}[1]{>{\raggedleft\let\newline\\\arraybackslash\hspace{0pt}}m{#1}}
\usepackage[pagebackref=true,breaklinks=true,letterpaper=true,colorlinks,bookmarks=false]{hyperref}

\def\eg{\emph{e.g}\onedot, }
\def\ie{\emph{i.e}\onedot, }

\usepackage[capitalize]{cleveref}
\crefname{section}{Sec.}{Secs.}
\Crefname{section}{Section}{Sections}
\Crefname{table}{Table}{Tables}
\crefname{table}{Tab.}{Tabs.}

\iccvfinalcopy 
\def\iccvPaperID{8826} 

\ificcvfinal\fi

\begin{document}

\title{Speech2Lip: High-fidelity Speech to Lip Generation by Learning from a Short Video}
\author{
Xiuzhe Wu$^{1}$,
~Pengfei Hu$^{2}$,
~Yang Wu$^{3,4}$,
~Xiaoyang Lyu$^{1}$,
~Yan-Pei Cao$^{3}$,
~Ying Shan$^{3}$,
~Wenming Yang$^{2}$,
~Zhongqian Sun$^{4}$,
~Xiaojuan Qi$^{1}$,
\vspace{0.8em}\\
$^{1}$The University of Hong Kong, $^{2}$  Tsinghua University,  $^{3}$  ARC Lab, Tencent PCG,  $^{4}$ Tencent AI Lab
}

\maketitle
\ificcvfinal\thispagestyle{empty}\fi

\begin{abstract}
Synthesizing realistic videos according to a given speech is still an open challenge. Previous works have been plagued by issues such as inaccurate lip shape generation and poor image quality. The key reason is that only motions and appearances on limited facial areas (e.g., lip area) are mainly driven by the input speech. Therefore, directly learning a mapping function from speech to the entire head image is prone to ambiguity, particularly when using a short video for training. We thus propose a decomposition-synthesis-composition framework named Speech to Lip (\textbf{Speech2Lip}) that disentangles speech-sensitive and speech-insensitive motion/appearance to facilitate effective learning from limited training data, resulting in the generation of natural-looking videos. First, given a fixed head pose (i.e., canonical space), we present a speech-driven implicit model for lip image generation which concentrates on learning speech-sensitive motion and appearance. Next, to model the major speech-insensitive motion (i.e., head movement), we introduce a geometry-aware mutual explicit mapping (GAMEM) module that establishes geometric mappings between different head poses. This allows us to paste generated lip images at the canonical space onto head images with arbitrary poses and synthesize talking videos with natural head movements. In addition, a Blend-Net and a contrastive sync loss are introduced to enhance the overall synthesis performance. Quantitative and qualitative results on three benchmarks demonstrate that our model can be trained by a video of just a few minutes in length and achieve state-of-the-art performance in both visual quality and speech-visual synchronization. Code: \href{https://github.com/CVMI-Lab/Speech2Lip}{https://github.com/CVMI-Lab/Speech2Lip}.

\end{abstract}

\section{Introduction}
\begin{figure}[!t]
    \centering
    \includegraphics[width=\linewidth]{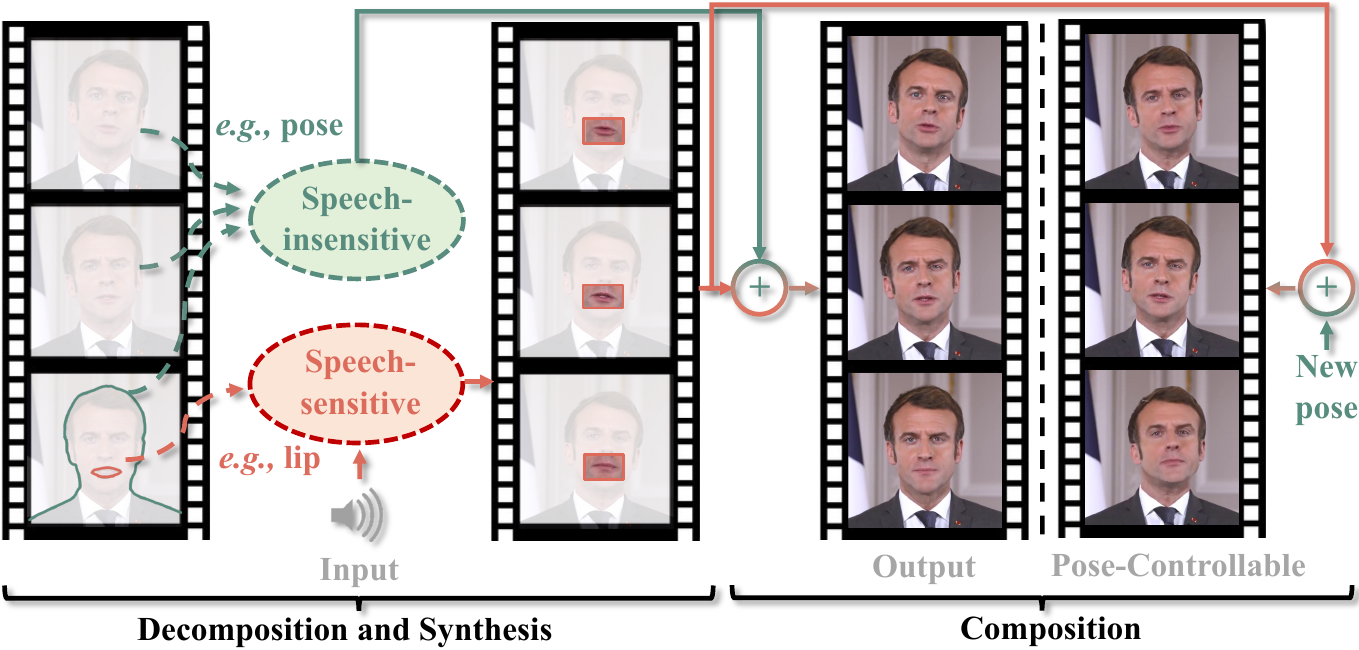}
    \caption{Given a speech as input, our model generates high-quality talking-head videos and supports pose-controllable synthesis. The decomposition and synthesis modules make learning from a short video more effective and the composition module enables us to synthesize high-fidelity videos.}
    \label{fig:teaser}
\end{figure}
Learning from a short video to generate personalized audio-synchronized talking videos driven by a speech is of great importance to various applications, for instance, digital human animation, video dubbing, and UGC video creation. However, synthesizing high-fidelity videos from speech for a desired speaker remains a challenging task.

The first challenge arises from \textit{complicated motion patterns}. 
Although speech majorly influences lip areas ({\ie} speech-sensitive areas), 
lip movements are often accompanied by other motions, such as global head movements, which greatly impact lip shapes and appearances. Thus, directly synthesizing a whole image from speech often leads to inaccurate lip synthesis. Second, existing methods still struggle to satisfy \textit{appearance fidelity} requirements, which include both identity preservation (speaker-specific) and high-quality detail generation, such as clear details of teeth, tongue, and eye-blinking \cite{chung2017you, zhou2019talking, prajwal2020lip, zhou2020makelttalk, chen2019hierarchical}. Third, it's difficult to \textit{acquire videos longer than 10 hours} for a  speaker which is yet required by conventional methods \cite{kumar2017obamanet, suwajanakorn2017synthesizing}.

To tackle the aforementioned challenges, existing attempts can be coarsely categorized into two lines of research: speaker-independent and speaker-specific methods. The first line often exploits GANs~\cite{goodfellow2014generative} that need to be trained on large-scale multi-person datasets.  
However, GAN-based models~\cite{chen2020comprises,chen2019hierarchical,chung2017you,prajwal2020lip,vougioukas2020realistic,wiles2018x2face,zhou2019talking,zhou2021pose} usually synthesize low-resolution images and unnatural motions (\ie background movements). They thus hardly meet the appearance fidelity requirement in terms of sharpness and fine appearance details. Moreover, preserving the identity of the speaker remains a challenging task \cite{chen2019hierarchical,zhou2021pose}. 

For the consideration of high-fidelity and identity preservation, another strategy focuses on learning from a specific speaker. Although attaining high-fidelity results, early works~\cite{kumar2017obamanet, suwajanakorn2017synthesizing} often require {several hours of video footage from a speaker} for training, hindering their practical applicability. 
Recently, NeRF~\cite{mildenhall2020nerf} has emerged as a promising approach for generating high-fidelity videos, which succeeds in learning from a short video of just a few minutes and having the potential to generate high-fidelity results~\cite{guo2021ad,liu2022semantic,shen2022dfrf}. 
Nevertheless, the models still struggle with appearance and motion ambiguity issues because they model speech-sensitive motions/appearances together with other facial areas less correlated to the given speech. 
This issue becomes more severe when training data is limited since no extra information can be exploited to avoid interference from signals that are not correlated with the speech.
As a result, they tend to generate lip sequences that do not synchronize well with the speech and produce blurry images (see {Figure} \ref{fig:qualitative_res} and Table \ref{tab:quantitative}). 
Therefore, reducing the complexity of modeling motions is critical to enable effective learning from a short video and synthesizing high-quality videos for a specific speaker.  

We thus design a preliminary experiment to identify that motion and appearance of lip areas have a strong correlation with speech, while head motion and other facial areas are less related to speech. Motivated by the observation, we propose decomposing speech-insensitive motion/appearance from speech-sensitive one, {and synthesizing them separately before composing them into a new talking video that aligns with the given speech (Figure \ref{fig:teaser}).} 
Toward this goal, we present a decomposition-synthesis-composition framework named Speech to Lip (\textbf{Speech2Lip}).
{In the decomposition stage, we introduce a speech-driven implicit model that generates high-fidelity synced lip sequences in a fixed view (\ie canonical view). To model 3D head motion effectively, we design a Geometry-Aware Mutual Explicit Mapping (GAMEM) module that estimates explicit geometric mappings between an arbitrary observed view and the canonical view. GAMEM also includes a jointly optimized canonical-view full-head depth map, which enables the model to be 3D-aware and supports controllable synthesis driven by new head poses. In the composition stage, GAMEM allows us to flexibly paste the synthesized canonical-view lips onto an arbitrary observed view to obtain natural synchronized talking videos. To improve the synthesis and synchronization qualities {after composition}, we incorporate a blending network (\ie Blend-Net) to refine the results and a contrastive sync loss to facilitate learning from a short video for generating synchronized talking videos.}

Our major contributions are summarized as follows:
\begin{itemize}
    \item [1)] 
        {We introduce a novel framework that disentangles speech-sensitive and speech-insensitive motion/appearance in high-fidelity video synthesis. By separating these components, the framework can effectively learn from limited training data.}
    \vspace{-0.05in}\item [2)]
        {The proposed speech-driven implicit model synthesizes speech-sensitive contents and the GAMEM flexibly combines them with given speech-insensitive areas to generate synchronized talking heads with natural movements and support pose-controllable synthesis.}
    \vspace{-0.05in} \item [3)]
        {Both qualitative and quantitative experimental results demonstrate the superiority of our method over the state-of-the-art speaker-specific methods.}
\end{itemize}

\section{Related Work}
\vspace{0.05in}\noindent{\textbf{Speech-Driven Talking Face Synthesis}{.}}
Video synthesis from {speech} is a long-standing problem. {Recent works can be generally divided into two categories: speaker-specific \cite{kumar2017obamanet, suwajanakorn2017synthesizing, lu2021live, guo2021ad, yao2022dfa, liu2022semantic, shen2022dfrf} and speaker-independent \cite{chung2017you, zhou2019talking, prajwal2020lip, zhou2020makelttalk, chen2019hierarchical, chen2020talking, zhang2021flow, wang2021audio2head, wang2022one}.}
In the first track, earlier works have succeeded in obtaining realistic visual results for a target person \cite{kumar2017obamanet, suwajanakorn2017synthesizing} but required hours of video belonging to one specific speaker. 
Therefore, many efforts \cite{lu2021live, guo2021ad, yao2022dfa, liu2022semantic, shen2022dfrf} have been made to {train} the model with a shorter talking video (3-5 minutes).
However, they utilize speech to forecast overall motions including speech-insensitive ones, failing to learn accurate lip shapes and appearances. The other line (\ie speaker-independent method) aims to build a universal model for all identities \cite{chung2017you, zhou2019talking, prajwal2020lip, zhou2020makelttalk, chen2019hierarchical}. 
The end-to-end pipeline \cite{chung2017you, zhou2019talking} is usually developed on GAN \cite{goodfellow2014generative}.
To boost performance, {Richard \etal \cite{richard2021meshtalk} only synthesize the mesh of limited facial areas,} Prajwal \etal \cite{prajwal2020lip} present an extra lip-sync discriminator, Zhou \etal \cite{zhou2021pose} provide additional head poses as inputs, and some other models \cite{chen2019hierarchical, das2020speech, zhou2020makelttalk, thies2020neural, song2022everybody} leverage the intermediate representation (\eg 2D facial landmarks \cite{chen2019hierarchical, das2020speech, zhou2020makelttalk} or expression parameters \cite{thies2020neural, song2022everybody}).
Nevertheless, they often suffer from low video quality, difficult identity preservation, and abnormal head motion generation.

\paragraph{\textbf{Implicit Representation based Talking Head Methods}.} 
Recently, implicit representations have shown high modeling capabilities in multiple tasks~\cite{park2019deepsdf,mildenhall2020nerf,oechsle2021unisurf,wang2021neus}. Among them, NeRF~\cite{mildenhall2020nerf} obtains extraordinary performance in novel view synthesis by training on only hundreds of images. Guo \etal \cite{guo2021ad} first apply it in the speaker-specific talking head synthesis task to enable learning from a short video. However, they utilize two NeRFs to model the head and torso, resulting in the head-torso separation phenomenon. Liu \etal \cite{liu2022semantic} solve it by leveraging one unified NeRF with two semantic-aware modules. Shen \etal \cite{shen2022dfrf} incorporate 2D image features as additional inputs to further reduce training data requirements (\ie videos with only 10-15 seconds). Regardless, these methods only ensure visual quality, and simply use speech to drive all the complex motions, leading to the out-of-sync problem. 

\section{Empirical Study and Motivations}
\begin{figure}[htbp]
    \centering
    \includegraphics[width=0.8\linewidth]{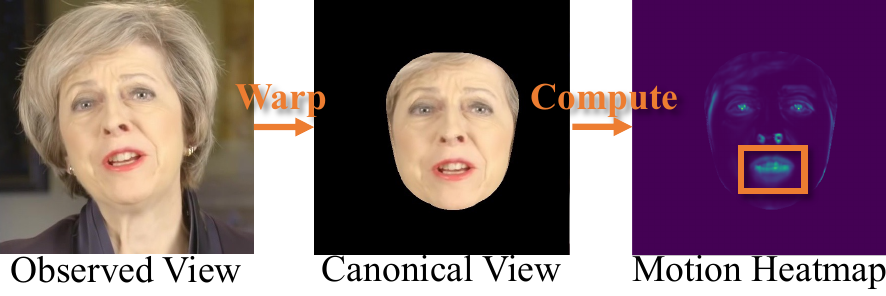}
    \vspace{-0.2cm}
    \caption{Speech-sensitive Motion Heatmap.} 
    \label{fig:heatmap}
    \vspace{-0.7cm}
\end{figure}

Our approach is motivated by the observation that only limited facial areas are highly correlated to speech. To verify this, we conduct a preliminary experiment to determine which areas are most speech-sensitive. Specifically, we apply warping to all captured images with varying head poses (refer to observed views) using the 3D Morphable Model (3DMM)~\cite{blanz1999morphable} and bring them to a fixed head pose (known as the canonical view). We then compute a motion heatmap in the canonical view to determine the areas most responsive to speech. As depicted in Figure \ref{fig:heatmap}, the elimination of head motion results in most of the motions around the lip regions displaying high sensitivity to the input speech. Additional examples can be found in the supplementary file. For more details on the warping process, please refer to Sec. \ref{sec:mapping}.

\begin{figure*}[!t]
    \centering
    \includegraphics[width=\linewidth]{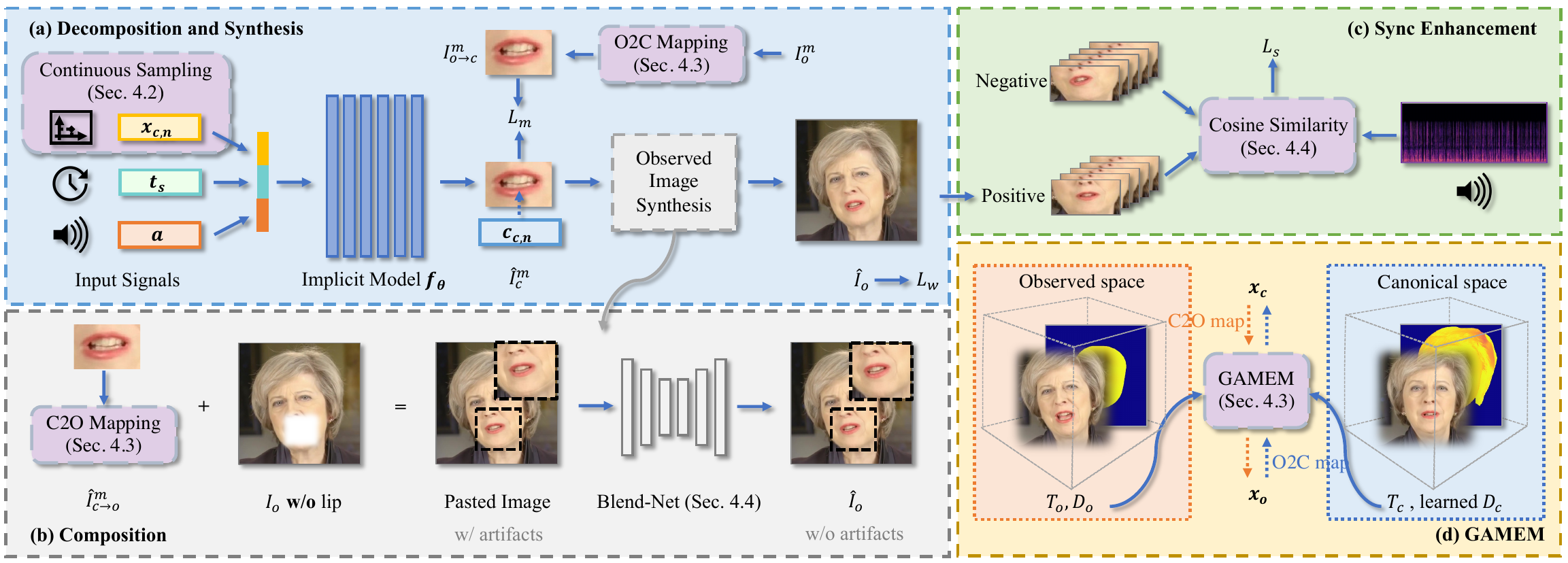}
    \caption{The overall framework of Speech2Lip. {Our framework decomposes speech-sensitive and speech-insensitive motions/appearances first, models them individually (the major part of (a)), and composes the ultimate output image (b). Besides, the synchronization performance enhancement module and the GAMEM are illustrated in (c) and (d), respectively.}
    The inputs include continuous pixel coordinates, speech audio signals, and timestamps. 
    The speech-driven implicit model will generate speech-sensitive canonical-view lip images, which will be further transformed into observed space to composite the eventual output image. A full-head depth map is learned along with the training process, supporting pose-controllable synthesis.}
    \label{fig:framework}
\end{figure*}
\section{Method}
\subsection{Overview} 
Figure ~\ref{fig:framework} provides an overview of our model. 
We define each frame captured in the observed view with its own head motion as the ``observed space".
To disentangle speech-sensitive and speech-insensitive motions (\ie head motions), we randomly select one observed space to serve as the ``canonical space". By doing so, we are able to align all observed spaces with the canonical space, thereby eliminating the effects of head motions. Note that once selected, the canonical space is fixed during training and inference. To model speech-sensitive motions, we propose a speech-driven implicit model to generate lip images in the canonical space without head motion effect (Sec. \ref{sec:model}). Thus, the generated lip image is canonical-view. {A} Geometry-Aware Mutual Explicit Mapping (GAMEM) module {is further proposed} to model the speech-insensitive head motions without speech effect.  {Next}, we project {the lip images in canonical space generated by the above implicit model} to the observed space based on GAMEM so that the synthesized lip image can be aligned and composed with any arbitrary observed frame, giving the model the flexibility for diversified composition (Sec. \ref{sec:mapping}).
Finally, a blending network (\ie Blend-Net) and a contrastive sync loss are presented to improve the quality of final synthetic images $\hat I_o$ {(Sec. \ref{sec:blending})}.

\subsection{Disentangled and Synced Implicit Modeling}\label{sec:model}
The central part of Speech2Lip is the disentangled and synced {speech-driven} implicit generator (Figure~\ref{fig:framework}(a)), 
which focuses on generating 2D lip appearances that are synchronized with cross-modal speech. By employing motion and appearance disentanglement, the generator only retains speech-sensitive motions and appearances. A lip-syncing contrastive loss is utilized (Figure~\ref{fig:framework}(c)) so that it can generate synced high-fidelity mouth appearance by learning from very short video data. All these together make the implicit modeling disentangled and synced.

\vspace{0.05in}\noindent\textbf{Speech-driven Implicit Model}
To achieve high-fidelity canonical lip image generation, we utilize a speech-driven implicit model. An implicit model is defined as a function that maps the coordinate signal to another signal \cite{sitzmann2020implicit}, {\eg} color. The input coordinates of the implicit model are defined in a continuous space (\ie the real field) so that it helps exploit the inherent relationship between spatially adjacent locations. Also, it can be further mapped into the frequency domain and benefit the learning of high-frequency information \cite{mildenhall2020nerf}. {Besides,} by learning a dedicated model for a specific scenario, implicit models also gain a strong ability to generate high-quality desired outputs. In our model, color information $\boldsymbol{\hat c_{c,n}}=(r, g, b)$ is regarded as our output signal (appearance color), and the canonical implicit function $\boldsymbol{f_{\theta}}$ can be defined as
\begin{equation}
	\boldsymbol{\hat c_{c,n}} = \boldsymbol{f_{\theta}}(\boldsymbol{x_{c,n}}, \boldsymbol{a}, \boldsymbol{t}_s),
	\label{eq:implicit_moxdel} 
\end{equation}
where $\boldsymbol{x_{c,n}}$ is the continuous 2D pixel coordinate vector (u,v) in the canonical space. $\boldsymbol{a}\in\mathbb{R}^{64}$ is the feature vector of the input speech at the concerned moment and $\boldsymbol{t_s}$ represents {the} timestamp {information, which} is utilized to enhance {the} temporal consistency between adjacent frames.

\begin{figure}[t]
    \centering    
    \includegraphics[width=0.6\linewidth]{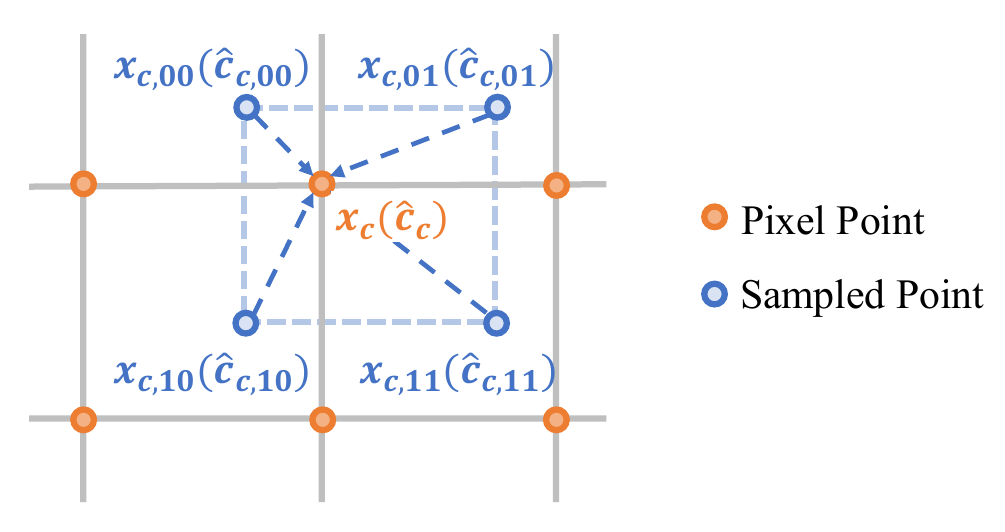}
    \caption{Continuous Sampling strategy. {This strategy can generate supervision signals for sampled positions at different resolutions in the training time.}} 
    \label{fig:local}
    \vspace{-0.2in}
\end{figure}

\vspace{0.05in} \noindent{\textbf{{Continuous Sampling}}}
The next question that arises is how to obtain the corresponding supervision signal to train the implicit model, which poses a challenge as the coordinates of the pixels are always integers, while we define the coordinates in a continuous space for more expressive features. {To overcome this challenge, we adopt the approach developed in~\cite{chen2021learning}, which enables us to sample and generate corresponding supervisions in the continuous coordinate space.} Particularly, for each pixel $\boldsymbol{x_c}$, a rectangle with an arbitrary shape is randomly sampled, the four corner points (\ie $\boldsymbol{x_{c,n}}$, $\boldsymbol{n}\in\{00,01,10,11\}$) of which are regarded as our inputs. A weighted averaged value $\boldsymbol{\hat c_c}$ is then calculated as
\begin{equation}
        \boldsymbol{\hat c_c} = \sum_{n\in\{00,01,10,11\}} \frac{S_n}{S}\cdot\boldsymbol{\hat c_{c,n}}, 
\end{equation}
where $\boldsymbol{\hat c_{c,n}}$ are the output appearance values of the sampled points. As illustrated in Figure~\ref{fig:local}, the blue-dashed-line rectangle represents the area $S$, which is divided into four sub-rectangles by gray lines, each with an area $S_n$. By leveraging this strategy, we can generate supervision for points with continuous positions, which enables us to fully exploit the advantages of the implicit model and achieve high-fidelity visual results.

\subsection{Geometry-Aware Mutual Explicit Mapping}\label{sec:mapping}
Once we have obtained the speech-sensitive lip sequences in the canonical view, we can construct the eventually synthesized image $\hat I_o$ by assigning the synthesized pixels to a location in the observed space. However, since the assigning process is speech-insensitive and only relies on the geometry information, we introduce the GAMEM module, illustrated in Figure \ref{fig:framework}, to explicitly model it. The GAMEM module comprises Canonical-to-Observed (C2O), Observed-to-Canonical (O2C) Mapping, and a learnable full-head depth map, which supports various applications, including pose-controllable image synthesis.

\vspace{0.05in}\noindent{{\textbf{C2O/O2C Mapping}}} 
{These mappings represent how the images transform between two spaces ({\ie} canonical space and observed space). As depicted in Figure \ref{fig:framework}, O2C Mapping aims at creating supervisions $I_{o\rightarrow c}^m$ for training the lip generator model $\boldsymbol{f_{\theta}}$  and C2O Mapping is utilized to warp the generated lip image into the observed view to produce $\hat{I}_{c\rightarrow o}^m$ for further composition. As they follow similar principles, we take the O2C Mapping as an example to illustrate the process. The input of O2C Mapping contains $I_o$ and $I_c$, and the output are pixel correspondences that map pixel coordinates in observed space to the canonical space. To achieve it, we use the 3D Morphable Model~(3DMM)~\cite{blanz1999morphable} to calculate the overall camera intrinsic matrix $K$ and rough face geometry $G$. The head poses $T_o\in\mathbb{R}^{4\times4}$ and $T_c\in\mathbb{R}^{4\times4}$, which include a rotation matrix and a translation vector, are also estimated. And the relative head pose $T_{c \rightarrow o}\in\mathbb{R}^{4\times4}$ between two spaces can be computed as
\begin{equation}
    T_{c \rightarrow o} = T_o \times T^{-1}_{c}.
    \label{eq:rel_pose}
\end{equation}

Furthermore, \textit{face} depth maps $D_o$ and $D_c$ for $I_o$ and $I_c$ obtained from 3DMM can be further interpolated based on $G$, and the corresponding head poses. 
Noticing that 3DMM can only be aware of the face area instead of the head area, but the information about the rest part of the head (\eg forehead, ears, eyes, mouth) is lacking. Hence we optimize the $D_c$ along with our model {to complete the missing depth region}. 
If $p_o$ and $p_c$ are the corresponding 2D homogeneous pixel grid coordinates on $I_o$ and $I_c$, the geometric relationships can be explicitly formulated as 
\begin{equation}
    D_op_o = KT_{c \rightarrow o}D_cK^{-1}p_c,
\end{equation}
so the position mapping from canonical space to observed space can be easily obtained as
\begin{equation}
    \boldsymbol{F_{c\rightarrow o}}(T_{c \rightarrow o}, D_c, p_c) = p_o.
    \label{eq:mapping}
\end{equation}

Then, we can warp the observed ground-truth lip images $I^m_o$ into the canonical space by \textit{backward warping}, {denoted by $I^m_{o\rightarrow c}$}, to guide the learning of the lip generation model. The C2O Mapping can be similarly defined by $T_{o \rightarrow c}$ and $D_o$. With it, we can warp the generated canonical lip image $\hat I^m_{c}$ into the observed space by \textit{backward warping}, denoted by $\hat{I}_{c \rightarrow o}^m$, to compose with the given observed image $I_{o}$ for synthesizing a talking face. 

\vspace{0.05in} \noindent \textbf{Pose Controllable Synthesis} 
Thanks to the simultaneously learned complete depth map $D_c$, pose-controllable full-head image synthesis according to users' requirements is also supported in our model. Specifically, we employ $D_c$ to determine location correspondences based on Eq. \ref{eq:rel_pose} and Eq. \ref{eq:mapping} by altering $T_o$. There is a slight difference in this situation, which is to use the \textit{forward warping} strategy as the ground-truth depths at new given poses are missing, which will produce black holes. To mitigate this issue, we propose a data augmentation method in training by randomly adding black holes with a ratio of 50\% to the image to let the Blend-Net learn to fill these areas.  

\subsection{{Overall Refinement}}
\vspace{0.05in} \noindent \textbf{Image Blending}
\label{sec:blending}
With $\hat{I}_{c \rightarrow o}^m$, we can integrate it by directly pasting it to the original observed frame based on 2D lip landmarks~\cite{bulat2017far}. However, the paste operation often results in mismatching artifacts in the boundary area, and the image artifacts after data augmentation should also be modified. 
Therefore, we propose a Blend-Net for blending.
The Blend-Net takes the pasted image after the paste operation as input to synthesize harmonized final output $\hat{I}_o$. Since the target is fusion and amending instead of generation, we predict the pixel residual, which is further added back to the input image to composite the final output image. {Detailed network structure is illustrated in the supplementary file.}

\vspace{0.05in} \noindent \textbf{{Synchronization Enhancement.}}
\label{sec:sync_loss}
To further improve the synchronization performance, we introduce a pre-trained sync expert network { to boost the model's performance in synchronization inspiring by} \cite{prajwal2020lip}. 
The sync expert network consists of two pre-trained encoders, which extract features of image and speech audio within a sliding window, denoted by $\boldsymbol{i}$ and $\boldsymbol{a}$ respectively. Different from \cite{prajwal2020lip}, we only have a short video for training. Therefore, the unsynced speech-image pairs are also exploited to construct the contrastive loss which helps avoid falling into a trivial solution. Its effectiveness is also verified in our experiments. The distance of synced speech-image pairs is desired to be closer while that of the unsynced speech-image pairs should be farther. Thus, we define a contrastive sync loss $\mathcal{L}_{s}$ based on the metric which is widely used in contrastive learning:
\begin{equation}
    \mathcal{L}_{s} = y\cdot(1-\mathrm{cos}(\boldsymbol{i}, \boldsymbol{a}))+ (1-y)\cdot \mathrm{max}(0, \mathrm{cos}(\boldsymbol{i}, \boldsymbol{a})),
    \label{eq:loss_sync} 
\end{equation}
where $\mathrm{cos}$ is the cosine similarity metric. $y=1$ and $y=0$ represent positive and negative speech-image pairs, respectively. For the input speech, the positive images are the matched images at the same timestamp while the negative images are randomly chosen at some timestamps else.

\subsection{Training Objectives}
\label{sec:training}
Our whole pipeline is trained in a self-supervised manner using the observed frames to provide the supervisory signals.
The overall loss includes a canonical mouth image reconstruction loss $\mathcal{L}_{m}$, a depth loss $\mathcal{L}_{d}$, a sync loss $\mathcal{L}_{s}$, {and a whole observed image reconstruction loss $\mathcal{L}_{w}$}. 

$\mathcal{L}_m$ measures the reconstruction error between the predicted lip image $\hat{I}_c^m$ and ground-truth lip image $I_{o\rightarrow c}^m$ as
\begin{equation}
    \mathcal{L}_{m} = \mathcal{L}_{p}(\hat I^{m}_c, I^{m}_{o\rightarrow c}) + \Vert\hat I^{m}_c-I^{m}_{o\rightarrow c}\Vert_2,\\\\
\end{equation}
where $I^{m}_{o\rightarrow c}$ is the warped ground truth lip image using O2C space mapping (see the supplementary file for more details) and the widely-used perceptual loss $\mathcal{L}_{p}$ is defined in \cite{zhang2018perceptual}. Similarly, overall reconstruction loss $\mathcal{L}_{w}$ is computed as
\begin{equation}
    \mathcal{L}_{w} = \mathcal{L}_{p}(\hat I_o, I_o) + \Vert\hat I_o-I_o\Vert_2.\\\\
\end{equation}

The canonical head depth map is initialized by the incomplete depth map computed by 3DMM and is trained with the help of the photometric loss as
\begin{equation}
    \mathcal{L}_{d} = \Vert\hat I_{o\rightarrow c}-I_c\Vert_2,\\\\
\end{equation}
where $\hat I_{o\rightarrow c}$ is the warped predicted image. With the contrastive sync loss (Sec. \ref{sec:sync_loss}), the overall loss function is
\begin{equation}
    \begin{aligned}
    \mathcal{L} = \omega_{m}\cdot\mathcal{L}_{m}+\omega_{w}\cdot\mathcal{L}_{w}+\omega_{d}\cdot\mathcal{L}_{d}+\omega_{s}\cdot\mathcal{L}_{s}.
    \label{eq:loss_overall} 
    \end{aligned}
\end{equation}
Implementation details are shown in the supplementary file.

\begin{table*}[!htbp]
\renewcommand{\arraystretch}{1.2}
\footnotesize
\centering
\scalebox{0.93}{
\setlength{\tabcolsep}{2mm}{
\begin{tabular}{l|c|ccccc|ccccc|c}
\toprule
\multirow{2}{*}{Method} & {Trained with} & \multicolumn{5}{c}{{Testset I}} & \multicolumn{5}{c}{{Testset II}} & \multicolumn{1}{c}{{Testset III}} \\
& {large extra data} & PSNR$\uparrow$ & SSIM$\uparrow$ & CPBD$\uparrow$ & $\mathrm{LMD}$$\downarrow$ & Sync$\uparrow$ & PSNR$\uparrow$ & SSIM$\uparrow$ & CPBD$\uparrow$ & $\mathrm{LMD}$$\downarrow$ & Sync$\uparrow$ & Sync$\uparrow$ \\
\midrule
{\textit{Ground Truth}} & {\textit{N/A}} & \textit{N/A} & \textit{1.000} & \textit{0.186} & \textit{0.000} & \textit{9.102}
& \textit{N/A} & \textit{1.000} & \textit{0.121} & \textit{0.000} & \textit{8.688} & \textit{4.877} \\
\midrule
 ATVG~\cite{chen2019hierarchical} & {Yes} & 28.452 & 0.818 & 0.019 & 5.423 & 5.813 & 28.051 & 0.668 & 0.003 & 4.203 & 6.224 & 3.571 \\
 MakeitTalk~\cite{zhou2020makelttalk} & {Yes} & 29.692 & 0.906 & 0.050 & 4.215 & 4.115 & 28.996 & 0.813 & 0.061 & 4.463 & 5.559 & 2.320 \\
 Wav2Lip~\cite{prajwal2020lip} & {Yes} & \textbf{31.557} & \textbf{0.980} & \textbf{0.115} & \textbf{3.053} & \underline{\textbf{10.031}} & \textbf{31.793} & \textbf{0.956} & \textbf{0.065} & \textbf{3.415} & \underline{\textbf{9.936}} & \underline{\textbf{5.809}} \\
 PC-AVS~\cite{zhou2021pose} & {Yes} & 29.072 & 0.880 & 0.040 & 4.595 & 9.258 & 28.359 & 0.734 & 0.046 & 4.305 & 8.586 & 5.206 \\
 \hline
 LSP~\cite{lu2021live} & {No} & 29.515 & 0.900 & 0.098 & 3.174 & 5.377 & 28.895 & 0.776 & 0.117 & 4.972 & 6.811 & 3.046 \\
 AD-NeRF~\cite{guo2021ad} & {No} & 32.223 & 0.954 & 0.051 & 2.989 & 6.042 & 30.885 & 0.909 & 0.055 & 3.210 & 5.910 & 3.285 \\
 DFRF~\cite{shen2022dfrf} & {No} & 33.292 & 0.974 & 0.094 & 3.079 & 5.252 & 31.419 & 0.944 & 0.124 & 3.139 & 5.552 & 2.879 \\
 \textbf{Speech2Lip} & {No} & \underline{\textbf{34.815}} & \underline{\textbf{0.987}} & \underline{\textbf{0.224}} & \underline{\textbf{2.976}} & {\textbf{7.771}} & \underline{\textbf{33.197}} & \underline{\textbf{0.962}} & \underline{\textbf{0.125}} & \underline{\textbf{3.082}} & {\textbf{7.370}} & {\textbf{4.379}} \\
\bottomrule
\end{tabular}
}
}
\caption{Quantitative results compared with the SOTA methods. {Image quality assessment metrics (\ie PSNR, SSIM, and CPBD) are computed within \textbf{mouth region}.} Algorithms are categorized into two classes based on the training datasets for a fair comparison. The first ones are trained in large public datasets while the other ones are trained using only 3-5min videos. {The best results of each class are in \textbf{bold} and overall best results are with \underline{\textbf{underlines}}}. }
\label{tab:quantitative}
\end{table*}

\section{Experiments}
\subsection{Datasets and metrics}
\paragraph{\textbf{Datasets}{.}}
We evaluate our algorithm and other methods on three datasets belonging to three specific speakers~({Testset I}, {Testset II} and {Testset III}) as recent speaker-specific methods \cite{guo2021ad, lu2021live} usually do.
Training videos are collected from \cite{lu2021live} as they are publicly available.
All video sequences are resampled to 25 FPS and split into {training part} and {test part} with a ratio of 90\%/10\%. The training splits of Testset I ($500\times500$) and Testset II ($624\times624$) are employed to train the models and then same-identity evaluations are conducted on the test splits. Testset III is utilized for cross-identity, -gender, and -language tests. We present more evaluation results in the supplementary file.

\vspace{0.05in}\noindent{\textbf{Evaluation Metrics.}}
Since both face generation and lip generation models are contained in our quantitative evaluations, we conduct the image quality assessment just around the \textbf{mouth region} for a fair comparison. To evaluate the appearance quality, we first utilize image quality metrics Peak Signal-to-Noise Ratio~({PSNR}) and {SSIM}~\cite{wang2004image}.
{A no-reference objective image sharpness metric based on Cumulative Probability of Blur Detection~({CPBD})~\cite{narvekar2011no,narvekar2010improved,narvekar2009no} is further utilized to measure the overall perceptual image quality.}
For synchronization evaluation, Landmarks Distance (LMD) around the \textbf{mouth region} is exploited to compute the accuracy of lip shape following \cite{chen2018lip}. The confidence score of lip synchronization computed by a pre-trained SyncNet~\cite{chung2016out} is also applied to measure the synchronization performance, labeled by Sync. 

\subsection{Comparisons with State of the Arts}
\vspace{0.05in}\noindent{\textbf{{Same-identity Evaluation}}{.}}
For speaker-independent models, we conduct experiments using their officially released pre-trained models. 
Since AD-NeRF~\cite{guo2021ad} and DFRF~\cite{shen2022dfrf} are speaker-specific models, we retrain them using our training data for a fair comparison. DFRF aims at few-shot learning but its lip synchronization performance witnesses a significant drop when the number of training images decreases from ~5000 to 15. Hence, we still train DFRF on thousands of images. 
Also, we directly use LSP's~\cite{lu2021live} pre-trained models on {Testset I} and {Testset II} to conduct experiments (subject May for Testset I and Testset III, and subject Obama2 for Testset II). We cannot provide a quantitative comparison with SSP-NeRF~\cite{liu2022semantic} here, as neither codes nor pre-trained weights are available. Thus, for comparison, we extract the speech from their released demo and then present qualitative comparisons in the supplementary file. 

\begin{figure*}[t]
    \centering
    \includegraphics[width=\linewidth]{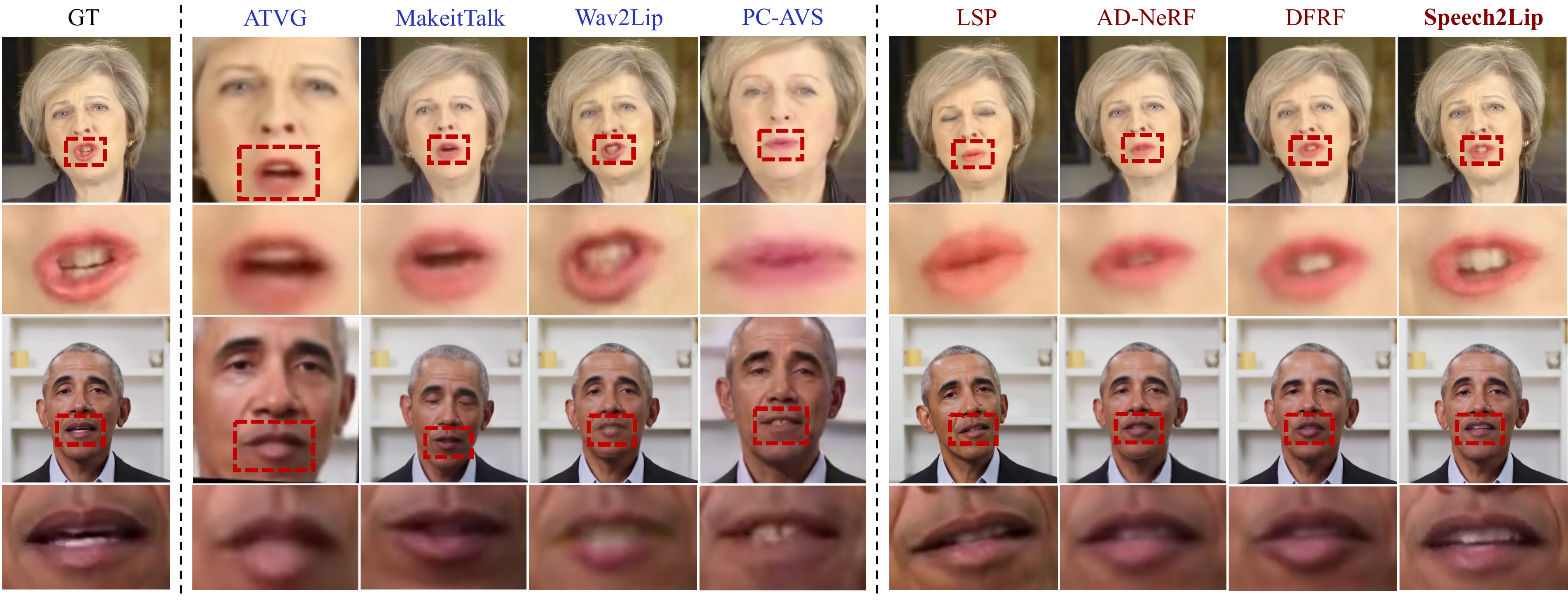}
    \vspace{-0.2in}
    \caption{Qualitative results compared with  SOTA methods. The Lip area is cropped based on the detected 2D landmarks for a clear comparison. {\textcolor[rgb]{0.161,0.204,0.710}{Speaker-independent} and \textcolor[rgb]{0.5,0,0}{speaker-specific} models are remarked by different colors.}} 
    \vspace{-0.1in}
    \label{fig:qualitative_res}
\end{figure*}

Quantitative results are shown in Table~\ref{tab:quantitative}. 
We divide {all the} models into two types based on the scale of the training dataset. The upper four algorithms (speaker-independent models) are trained using public video datasets with large amounts of identities and speech-image pairs~(\eg LRS2~\cite{Afouras18c, Chung17, Chung17a} and VoxCeleb2~\cite{chung2018voxceleb2}) while the latter four methods (speaker-specific models) only use 5-minute videos of a specific identity for training.
Speaker-independent models tend to perform well in synchronization {because they can learn lip motions well from various speech-visual pairs in training time.}
However, the cost is sacrificing the visual quality.
The metrics of image quality~(\eg PSNR, SSIM, and CPBD) are much lower than that of specific models. Wav2lip~\cite{prajwal2020lip} is an exception because it only generates lip images{, but the details still can not be produced well (Figure \ref{fig:qualitative_res}).}
In contrast, speaker-specific models usually have {much} better image qualities, which is favorable in practice. Our model outperforms all the SOTA algorithms in image quality and achieves the best synchronization performance among all the speaker-specific models, and the Sync score is also competitive when compared to those speaker-independent models. It is also worth mentioning that the Sync scores of Wav2lip~\cite{prajwal2020lip} and PC-AVS~\cite{zhou2021pose} are higher than that of the ground truth. The reason might be that the Sync score is sensitive to synchronization quality only when synchronization quality is within a range and a higher score above a threshold may not necessarily imply a higher synchronization quality. We design a toy experiment to verify it in the supplementary file and our user study can also demonstrate it.

\vspace{0.05in}\noindent{\textbf{{Cross-identity Evaluation}}{.}}
{Our model also has the ability to generate cross-identity, -gender, and -language results. 
{Since there is no ground truth image, we provide} the sync performance comparison on generating a video for the model trained on {Testset I}~(female, British) using another speaker's speech~(male, French). 
This setting is defined as {Testset III}, and results are shown in {the last column of} Table~\ref{tab:quantitative}. 
Our model still has significant superiority {and outperforms other speaker-specific models by a large margin.} }
{We also test the effect of unsynced speech-image pairs used in Eq.~\ref{eq:loss_sync} on {Testset III}. Without negative pairs, the Sync score decreases from 4.379 to 3.830, showing the effectiveness of our contrastive sync loss. }

\vspace{0.05in}\noindent{\textbf{Qualitative Results}{.}}
Qualitative comparisons of different algorithms are presented in {Figure}~\ref{fig:qualitative_res}. {We} can see that {speaker-independent methods} all suffer from low image quality. 
{Besides, ATVG and MakeitTalk have unnatural speech-insensitive movements (\eg head motion and background motion) and PC-AVS has an extra issue of struggling} with identity preservation.
{Speaker-specific approaches have much better visual quality but most of them} can hardly estimate {precise} lip {details} when there exist considerable significant lip motions. And fine details like teeth and tongue are not well modeled either. Besides, AD-NeRF sometimes causes head-torso separation {due to the adoption of} two separate NeRFs. In contrast, our method Speech2Lip performs well in all these situations, as the generated shapes are more accurate and the synthetic images are clearer and more realistic especially the lip area. More results can be found in the supplementary videos. 

\vspace{0.05in}\noindent{\textbf{Complexity Comparisons}.}
{To further demonstrate the superiority of our model, complexity and computational cost comparisons with speaker-specific models are conducted on Testset I. In Table ~\ref{tab:model_comparisons}, our model gains the best results with much less complexity and lower computational cost.}

\begin{table}[!htbp]
\renewcommand{\arraystretch}{1.2}
\centering
\scalebox{0.8}{
    \begin{tabular}{lcccc}
    \toprule 
    Methods & LSP & AD-NeRF & DFRF & \textbf{Ours} 
    \\\hline
    Model size (MB) $\downarrow$ & 500 & \underline{30} & \textbf{20} & \underline{30} \\
    Train time (hour) $\downarrow$ & \underline{38.5} & 80 & 60 & \textbf{30} \\
    Test speed (FPS) $\uparrow$ & \textbf{35} & 0.06 & 0.04 & \underline{18} \\
    PSNR (dB) $\uparrow$ & 29.52 & 32.22 & \underline{33.29} & \textbf{34.82} \\
    Sync $\uparrow$ & 5.38 & \underline{6.04} & 5.25 & \textbf{7.77} \\
    \bottomrule
    \end{tabular}
}
\caption{Speaker-specific model comparisons. The best results are in \textbf{bold} and the second best results are with \underline{underlines}.}
\label{tab:model_comparisons}
\end{table}

\vspace{0.05in}\noindent{\textbf{User Study}{.}}
To verify the {perceptual quality}, user studies {are conducted} on 16 {generated} video clips {covering both} {Testset I} and {{Testset II}. Videos are generated by our proposed} Speech2Lip and other SOTA algorithms, with algorithm names hidden and video order randomized. 
17 participants independently evaluated all the videos. For each video, each participant gave a Mean Opinion Score~(MOS) from 1-5 for each of three aspects: qualities of speech-visual synchronization, image fidelity, and image realness. Higher scores represent better quality. 
The overall results are shown in Table ~\ref{tab:user_study}. 
Speech2Lip not only achieves the highest MOS in image quality measurement (\eg 4.618 and 4.382) but also outperforms all the other algorithms in lip synchronization~(\eg 4.265), {exhibiting Speech2Lip has the most satisfactory overall video quality.} 

\begin{table*}[!htbp]
\renewcommand{\arraystretch}{1.2}
\footnotesize
\centering
\scalebox{1.0}{
\begin{tabular}{lccccccccc}
    \toprule
    Methods & ATVG & MakeitTalk & Wav2Lip & PC-AVS & LSP & AD-NeRF & DFRF & \textbf{Speech2Lip} 
    \\\hline
    Lip Synchronization & 1.500 & 1.853 & 4.191 & 3.059 & 3.118 & 3.088 & 2.853 & \textbf{4.265} \\
    Image Fidelity & 1.147 & 2.794 & 3.265 & 2.529 & 3.882 & 3.235 & 3.265 & \textbf{4.618} \\
    Image Realness & 1.118 & 2.088 & 3.765 & 2.265 & 3.000 & 2.324 & 2.971 & \textbf{4.382} \\
    \bottomrule
\end{tabular}
}
\caption{Detailed user study results compared with SOTA methods. The best results are in \textbf{bold}.}
\label{tab:user_study}
\end{table*}

\subsection{Ablation Study} 
\noindent{\textbf{Contributions of loss functions.}}
We first conduct an ablation study on loss functions. The evaluation metrics contain the PSNR around the lip area and the sync score. The PSNR of the whole image is also retained since practical applications attach more attention to the overall quality of the talking portrait scene.
Results are shown in Table ~\ref{tab:ablation_loss_3ch_lightBL}, demonstrating the loss functions all contribute to the increased performance on both visual quality and synchronization.

\begin{table}[t]
\renewcommand{\arraystretch}{1.2}
\footnotesize
\centering
\scalebox{1.0}{
    \setlength{\tabcolsep}{2.5mm}{
    \begin{tabular}{lccc}
        \toprule 
        Methods & $\mathrm{PSNR_{lip}}\uparrow$ &  $\mathrm{PSNR_{img}}\uparrow$ & Sync$\uparrow$ \\
        \hline 
        $\mathcal{L}_{w}$ & 34.104 & 36.947 & 6.313 \\
        $\mathcal{L}_{w}$ + $\mathcal{L}_{d}$ & 34.429 & 36.994 & 6.359 \\
        $\mathcal{L}_{w}$ + $\mathcal{L}_{d}$ + $\mathcal{L}_{m}$ & 34.694 & 37.193 & 7.194 \\
        \textbf{$\mathcal{L}_{w}$ + $\mathcal{L}_{d}$ + $\mathcal{L}_{m}$ + $\mathcal{L}_{s}$} & \textbf{34.815} & \textbf{37.245} & \textbf{7.771} \\
        \bottomrule
    \end{tabular}
    }
}
\caption{Ablation study results about loss function on {Testset I}.}
\label{tab:ablation_loss_3ch_lightBL}
\end{table}

\noindent{\textbf{Contributions of individual components.}} 
We also explore the benefits of our key components by removing each component to see how the performance changes in Table ~\ref{tab:ablation_module}. 
``\textbf{w/o} implicit model'' represents we employ an explicit lip generation model instead, and ``\textbf{w/o} time'' indicates timestamp is deleted from inputs. From the results, it can be concluded that each design plays an important role. Among them, the Blend-Net significantly improves video quality, and the design of implicit modeling schema is also critical.

\begin{table}[t]
\renewcommand{\arraystretch}{1.2}
\footnotesize
\centering
\scalebox{1.0}{
    \setlength{\tabcolsep}{2.mm}{
    \begin{tabular}{lccc}
        \toprule 
        Methods & $\mathrm{PSNR_{lip}}\uparrow$ &  $\mathrm{PSNR_{img}}\uparrow$ & Sync$\uparrow$ \\
        \hline
        \textbf{w/o} continuous sampling & 34.702 & 37.214 & 7.553 \\
         \textbf{w/o} implicit model & 34.250 & 37.058 & 6.790 \\
         \textbf{w/o} Blend-Net & 33.736 & 36.881 & 6.065 \\
         \textbf{w/o} time & 34.512 & 37.121 & 7.146 \\
         \textbf{Speech2Lip} & \textbf{34.815} & \textbf{37.245} & \textbf{7.771} \\
        \bottomrule
    \end{tabular}
    }
}
\caption{Ablation study results about network design on {Testset I}.}
\label{tab:ablation_module}
\end{table}

\subsection{Controllable Synthesis Results}
Thanks to the joint optimization and completion of a full-head depth map $D_c$ in GAMEM, our model supports synthesizing full-head images driven by novel head poses. Specifically, $D_c$ can be projected into any observed view to obtain $D_o$. $D_c$ and $T_o$ in Eq. \eqref{eq:rel_pose} are further replaced by $D_o$ and the new target head pose, respectively. Then, we adopt Eq. \eqref{eq:mapping} to calculate the correspondence. The correspondence is further exploited to warp {the synthesized image in the observed space} into a new target space. 
{It is noted that depth maps are used for coarse mapping between spaces while facial expression is more taken charge of by fine-grained contents from observed views.}
The pose-controllable novel view synthesis results are shown in {Figure~\ref{fig:novel_view_synthesis}}. 
{When testing with moderate pose changes, our model achieves comparable results as {AD-NeRF}.}

We note that existing methods with 3D modeling such as AD-NeRF also cannot generate good synthesis results if the evaluated poses deviate too much from the training data. The reason is that the training data widely used in speaker-specific models~\cite{lu2021live, guo2021ad, shen2022dfrf, liu2022semantic} only show the front view with limited pose variations, being naturally hard for 3D modeling. The 3D mesh is incomplete, and the surface is rough (Figure~\ref{fig:3d_vis_adnerf}).
In addition, the computational costs of 3D methods are extremely high (Table \ref{tab:model_comparisons}).

\begin{figure}[t]
\vspace{-0.1in}
    \centering
    \includegraphics[width=0.8\linewidth]{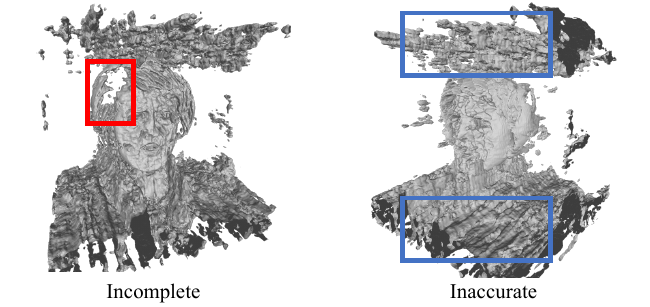}
    \vspace{-0.05in}
    \caption{3D Mesh of AD-NeRF on {Testset I}.}
    \vspace{-0.1in}
    \label{fig:3d_vis_adnerf}
\end{figure}

\begin{figure}[t]
    \centering
    \includegraphics[width=0.8\linewidth]{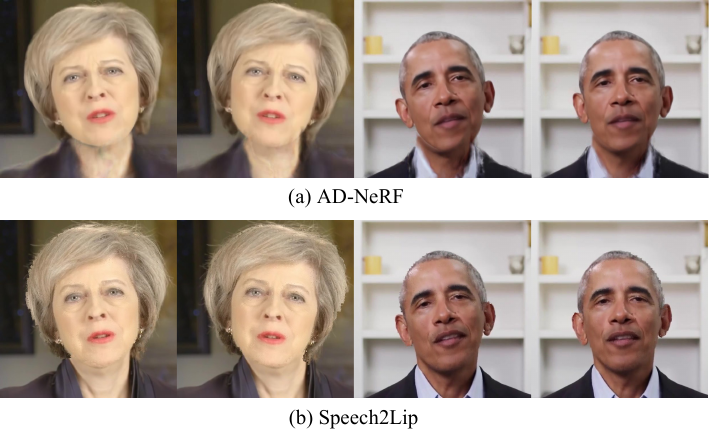}
    \vspace{-0.2cm}
    \caption{{Novel view synthesis capability of our model. {We rotate the head region with a random angle.}}}
    \label{fig:novel_view_synthesis}
    \vspace{-0.4cm}
\end{figure}

\section{Conclusion}
In this paper, we propose a novel decomposition-synthesis-composition framework called Speech2Lip for high-fidelity {talking head} video synthesis, which disentangles speech-sensitive and speech-insensitive {motions/appearances}. By {presenting} a synced speech-driven implicit model, a GAMEM module, a Blend-Net, and a contrastive sync loss, we can achieve satisfactory results with only a few minutes of {training} video. Our framework also supports pose-controllable synthesis. In the future, we plan to {study generating realistic expressions driven by speech and} explore combining our insights with advanced general image generation methods such as diffusion-based models for better generalizability. We hope that our work inspires more future research in this field and encourages the development of positive applications. However, we also urge caution to prevent any potential abuses. More discussions about limitations are described in the supplementary file.

\noindent \textbf{Acknowledgement} 
This work has been supported by the Research Fund from Tencent ARC lab.

{\small
\bibliographystyle{ieee_fullname}
\bibliography{egbib}
}

\end{document}